\documentclass[conference]{IEEEtran}
\IEEEoverridecommandlockouts
\usepackage{amsmath,amssymb,amsfonts}
\usepackage{amstext}
\usepackage{breqn}
\usepackage{algorithmic}
\usepackage{graphicx}
\usepackage{textcomp}
\usepackage{xcolor}
\usepackage{hyperref}
\usepackage{todonotes}
\usepackage{algorithm}
\usepackage{algorithmic}
\usepackage{bm}
\DeclareMathOperator*{\argmax}{arg\,max}
\def\BibTeX{{\rm B\kern-.05em{\sc i\kern-.025em b}\kern-.08em
    T\kern-.1667em\lower.7ex\hbox{E}\kern-.125emX}}
\begin{document}

\title{Continuous Episodic Control}

\author{\IEEEauthorblockN{Zhao Yang, Thomas. M. Moerland, Mike Preuss, Aske Plaat}
%\author{\IEEEauthorblockN{Anonymous submission}
\IEEEauthorblockA{\textit{Leiden Institute of Advanced Computer Science } \\
\textit{Leiden University}\\
z.yang@liacs.leidenuniv.nl}
}

\maketitle

\begin{abstract}
Non-parametric episodic memory can be used to quickly latch onto high-rewarded experience in reinforcement learning tasks. In contrast to parametric deep reinforcement learning approaches in which reward signals need to be back-propagated slowly, these methods only need to discover the solution once, and may then repeatedly solve the task. However, episodic control solutions are stored in discrete tables, and this approach has so far only been applied to discrete action space problems. Therefore, this paper introduces \textbf{C}ontinuous \textbf{E}pisodic \textbf{C}ontrol (CEC)\footnote{Code will be available after the authors notification.}, a novel non-parametric episodic memory algorithm for sequential decision making in problems with a continuous action space. Results on several sparse-reward continuous control environments show that our proposed method learns faster than state-of-the-art model-free RL and memory-augmented RL algorithms, while maintaining good long-run performance as well. In short, CEC can be a fast approach for learning in continuous control tasks.
%\footnote{Please see the anonymous code \href{https://www.dropbox.com/sh/jnbve0nlol24wie/AADz9s0sUOH99CiPApHvaP_Da?dl=0}{here}.}
\end{abstract}


\section{Introduction}
Deep reinforcement learning (RL) methods have recently demonstrated superhuman performance on a wide range of tasks, including Gran Turisma~\cite{wurman2022outracing}, StarCraft~\cite{vinyals2019grandmaster}, Go~\cite{silver2017mastering}, etc. However, in these methods, the weights of neural networks are slowly updated over time to match the target predictions based on the encountered reward signal. Therefore, even if the agent finds the solution to the problem, the learning still can be fairly slow, and the agent may not be able to solve the problem in subsequent episodes. This is one of the reasons state-of-the-art RL methods generally require many interactions with the environment and substantial computational resources~\cite{DBLP:journals/nature/MnihKSRVBGRFOPB15,DBLP:journals/nature/SilverHMGSDSAPL16}. Non-parametric episodic memory is introduced to directly store high-rewarded experiences, enabling the agent to quickly latch onto these experiences in reinforcement learning problems. Compared to parametric deep RL methods, these methods only need to solve the task once, because the solution is stored in memory and can be quickly retrieved. This is especially helpful for tasks where the agent only gets a final reward once it succeeds, i.e. sparse reward problems~\cite{ecoffet2021first}. Although parametric RL methods do perform well in the long run (due to its generalizability, optimality and ability of dealing with stochasticity), sometimes we prefer a quick and less computationally expensive solution to the problem~\cite{DBLP:journals/corr/BlundellUPLRLRW16}. The ability to quickly latch onto recent successful experiences is also clearly evident in nature, where scrub jays for example store food and directly remember the exact location using episodic memory~\cite{clayton1998episodic}.

Episodic memory is a term that originates from neuroscience~\cite{tulving1972episodic}, where it refers to memory that we can quickly recollect. In the context of RL, this concept has generally been implemented as a non-parametric (or semi-parametric) table that can be read from and written into rapidly. Information stored in the memory can then either be used directly for control (for action selection) or to generate training targets for parametric deep RL methods. When used for control (usually called episodic control), episodic memory stores state-action pairs and their corresponding episodic returns in memory and extracts the policy by taking actions with maximum returns. It is therefore conceptually similar to tabular RL methods~\cite{watkins1989learning} but with a different update rule, where stored values are directly replaced with larger encountered ones instead of being gradually updated towards targets. 

\begin{figure}[!tb]
    \centering
    \includegraphics[scale=0.18]{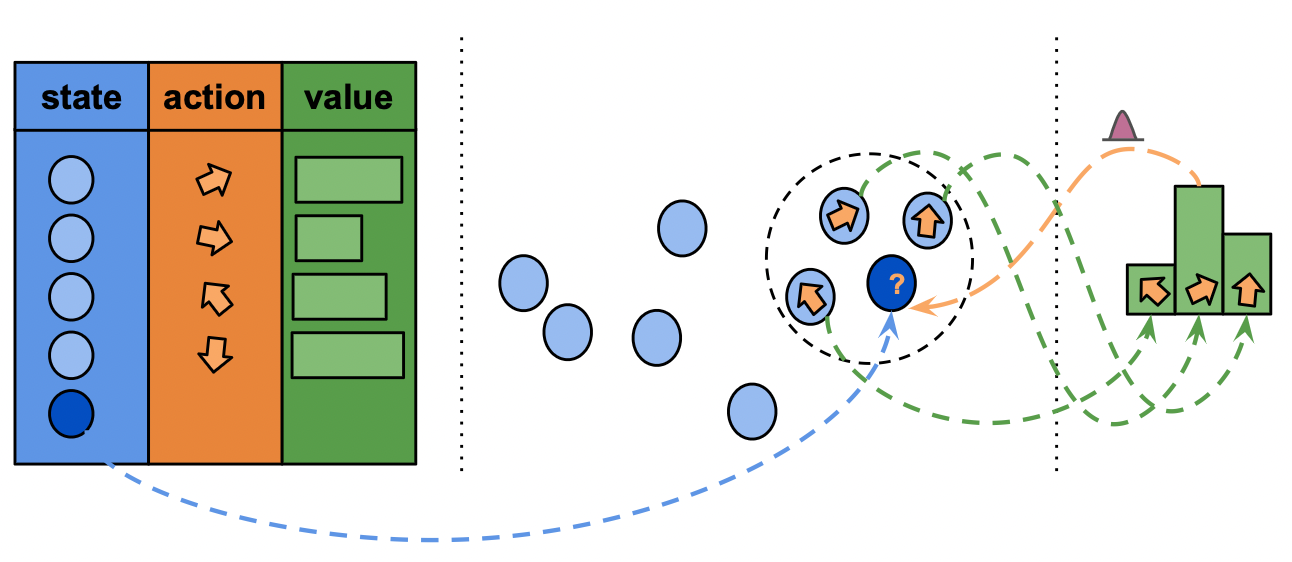}
    \caption{Conceptual illustration of the Continuous Episodic Control (CEC) algorithm. Left: We store a non-parametric table/buffer of state-action-value estimates obtained from previous episodes. Middle \& Right: Action selection for a novel query state (dark blue circle) is performed by collecting its nearest neighbours in the buffer (dashed circle), and then sampling one of the actions of these neighbours proportional to their value (right). After possibly injecting exploration noise the selected action for the query state is returned.}
    \label{fig:CEC}
\end{figure}

Since current episodic control methods maintain a value estimate for each state-action pair~\cite{DBLP:journals/corr/BlundellUPLRLRW16}, they cannot naturally deal with continuous action spaces. One solution to this problem is to combine episodic memory with deep RL methods, where the episodic memory part generates training targets for a value network that is used to act (value-based approaches~\cite{mnih2015human}) or to guide act (actor-critic approaches~\cite{lillicrap2015continuous}) in the environment. The drawback of this approach is that we not only have to maintain extra memory, but again rely on the relatively slow deep RL agent for action selection and thereby performance is affected. The natural question that arises is: can we directly use episodic memory for action selection in continuous control tasks, without relying on parametric RL methods?

This work therefore introduces \textbf{C}ontinuous \textbf{E}pisodic \textbf{C}ontrol (CEC), an algorithm that uses episodic memory directly for action selection in tasks with a continuous action space. The main idea is based on the principle of generalisation: similar states generally require similar actions to obtain high returns. However, given a new query state, we 1) may not have that exact state in our buffer, and 2) we definitely only have value estimates for a few points in the continuous space of possible actions. We solve these issues by 1) specifying the nearest neighbour search in state space and 2) selecting actions proportionally to their value estimate, injected with Gaussian noise per action. The overall process of CEC is illustrated in Fig.~\ref{fig:CEC}. 

Experiments on a range of continuous control tasks show that our method outperforms state-of-the-art model-free and memory-augmented RL methods in various continuous control problems. We also visualize how CEC manages to capture state-action values quickly and reliably in the beginning of the training, which results in steeper learning curves. In short, CEC seems to be a promising approach to learn quickly in continuous control tasks.
\section{Background}
In reinforcement learning~\cite{sutton2018reinforcement}, an agent interacts with the environment. This process is formed as a Markov Decision Process (MDP) defined as the tuple $M=\langle S,A,P,R,\gamma\rangle$, where $S$ is a set of states, $A$ is a set of actions the agent can take, $P$ specifies the transition dynamic of the environment, $R$ is the reward function, and $\gamma$ is the discount factor. At timestep $t$, the agent observes a state $s_t\in S$, and selects an action $a_t\in A$. The environment returns a next state $s_{t+1}\sim P(\cdot|s_t,a_t)$ and an associated reward $r_t=R(s_t,a_t,s_{t+1})$ after the action is executed. The agent selects actions according to a policy $\pi(\cdot|s_t)$ that maps a state to a distribution over actions. The goal is to learn a policy that can maximize the expected discounted return: $\mathbb{E}_{s_{t+1}\sim P(\cdot|s_t,a_t), a_t\sim \pi(\cdot|s_t), r_t\sim R(\cdot|s_t,a_t)}[\sum_{t=0}^{T}\gamma^{t}\cdot r_t]$, where $\gamma\in[0,1]$. 

Define the state-action value $Q(s_t,a_t)$ as the expected return: 
\begin{equation*}
    Q(s_t,a_t)=\mathbb{E}_{P, \pi, R}[\sum_{k=t}^{T}\gamma^k \cdot r_k]
\end{equation*}, when the agent starts from state $s_t$ and action $a_t$ at time-step $t$. Q-learning learns the optimal state-action value function by updating the current estimated value towards a target value (TD-target) following:
\begin{dmath*}
    \hat{Q}(s_t,a_t) \gets \hat{Q}(s_t,a_t) + \alpha \cdot[r_{t}+\gamma \cdot \max_{a'\in A} \hat{Q}(s_{t+1},a') - \hat{Q}(s_t,a_t)]
    \label{eq:q}
\end{dmath*}, where $\alpha$ is the learning rate, $\hat{Q}(s_t,a_t)$ is estimated state-action value.

In deep Q-learning, state-action value function $Q(s_t,a_t)$ is approximated by a neural network denoted by $Q_\phi(s_t,a_t)$. Then the function is updated by minimizing the  mean squared error $L(\phi, D)$ between the TD-target and the estimated value computed using samples from $D$:
\begin{dmath*}
    L(\phi, D) = \mathbb{E}_{(s_t,a_t,r_t,s_{t+1})\sim D} \Big(r_{t}+\gamma \cdot \max_{a'\in A}\hat{Q}_{\phi}(s_{t+1},a') - \hat{Q}_{\phi}(s_t,a_t)\Big )^2
    \label{eq:deepq}
\end{dmath*}
After the optimal state-action value function is learned, the optimal policy can be exacted by greedily taking actions that have highest state-action values:
\begin{equation}
a_t=\argmax_{a_t}Q(s_t,a_t). 
    \label{eq:q2policy}
\end{equation}
\section{Related Work}
In this section, we will discuss RL methods that use or are augmented with episodic memory. There are mainly two directions, one that directly uses episodic memory for control (`episodic control'), and another which uses data in the episodic memory to guide the learning update of parametric RL agents (`memory-augmented RL'). We will discuss both. 

\paragraph{Episodic control} Model-free episodic control (MFEC) ~\cite{DBLP:journals/corr/BlundellUPLRLRW16} uses a non-parametric table to store episodic return $G_t=\sum_{k=t}^{T}\gamma^{k-t}r_k$ for each state-action pair and only overwrites it when a higher return of the state-action pair is encountered:
\begin{dmath*}
    \hat{Q}_{em}(s_t,a_t) \gets \max[\hat{Q}_{em}(s_t,a_t), G_t]
\end{dmath*}
The optimal policy is extracted by greedily taking actions with the greatest values stored in the EM table, similar with Eq.~\ref{eq:q2policy}. It outperforms state-of-the-art model-free RL methods on some Atari games in the short run. Neural episodic control~\cite{pmlr-v70-pritzel17a} proposed to use a semi-tabular approach called differentiable neural dictionary (DND) which can store trainable representations which are updated during the training process. NEC selects action in the same way as MFEC. The slowly changing representations and quickly updating values lead to better performance than MFEC. To the best of our knowledge, no previous method has studied the use of episodic memory for action selection in continuous action spaces.

\paragraph{Memory-augmented RL} Instead of using non-parametric models directly for action selection, information stored can also be used to enhance learning of a deep RL agent. During updates of value-based deep RL methods~\cite{DBLP:journals/nature/MnihKSRVBGRFOPB15} or critic parts of actor-critic RL methods~\cite{DBLP:journals/corr/LillicrapHPHETS15}, the neural network is updated by minimizing the error between a TD target and the currently estimated $Q$ value. Therefore, values from the episodic memory can be used to form a new update target alongside the original TD target:

\begin{equation}
    L=\alpha \cdot (\hat{Q}-\hat{Q}_{original})^2 + \beta \cdot (\hat{Q}-\hat{Q}_{em})^2,
    \label{emrl}
\end{equation}

\noindent where $L$ is the loss/error to minimize, $\alpha$ and $\beta$ are hyper-parameters to balance the contribution of the original TD target $\hat{Q}_{original}$ and the episodic target $\hat{Q}_{em}$, and $\hat{Q}$ is the current Q value estimation. Episodic Memory Deep Q-networks (EMDQN)~\cite{10.5555/3304889.3304998}, Episodic Memory Actor-Critic (EMAC)~\cite{ijcai2021-365}, Model-based Episodic Control (MBEC)~\cite{DBLP:conf/nips/LeKATV21} and Generalizable Episodic Memory (GEM)~\cite{DBLP:conf/icml/HuYZRZ21} all use this approach, but in different ways. EMDQN combines values from the EM with 1-step TD target in deep Q learning using fixed $\alpha$ and $\beta$. EMAC combines values from the EM with 1-step TD target in DDPG for solving tasks with continuous action space. MBEC brings episodic memory into the model-based RL settings, and combines values from the EM with 1-step TD target using a learned weight instead of fixed $\alpha$ and $\beta$. GEM uses EM for implicit planning and performs the update with an $n$-step TD target~\cite{sutton2018reinforcement}, first taking $n$ steps based on the EM then bootstrapping from a Q-network. Aforementioned methods are all trying to distill information stored in EMs into parametric models. Although by doing so the performance of parametric models is enhanced, meanwhile it also loses the underlying benefit of EM which is fast reading and writing.

\begin{table}[!ht]
    \centering
    \caption{Overview of related memory-based RL approaches. None of these previous works uses episodic memory for action selection in problems with a continuous action space.}
    \begin{tabular}{c|c|c}
        Method & EM usage & Action space \\
        \hline
        MFEC & action selection & discrete \\
        NEC & action selection & discrete \\
        EMAC & target update & continuous \\
        EMDQN & target update & discrete \\
        GEM & target update & continuous \\
        MBEC & target update & discrete \\
        Neural Map & feature storage & discrete \\
        \hline
        \textbf{CEC (ours)} & \textbf{action selection} & \textbf{continuous} \\
    \end{tabular}
    \label{tab:overview}
\end{table}

Finally, note that there are various other methods, such as Neural Map ~\cite{DBLP:conf/iclr/ParisottoS18}, that learn to interact with an external (tabular) memory module to deal with partial observability, but these approaches do not store any value or policy estimates in the table, like episodic memory. A summary of related work is provided in Table \ref{tab:overview}. 

\section{Continuous Episodic Control (CEC)}
The principle of CEC is that actions the agent takes in similar states should also be similar. It is implemented by maintaining a table that contains a state-action-value tuple for each row. Each entry stores the cumulative discounted reward the agent received in an episode after taking the specific action in the specific state. To interact with the memory, CEC mainly has two functions: an \textbf{update} function that adds new encountered data into the memory or updates old data after an episode is terminated, and an \textbf{action selection} function that is used during the episode to select actions.

\paragraph{Update} We update the episodic memory after each collected episode. Intuitively, we want our episodic memory to have good coverage of the visited state space, for which we use the following mechanism. We first collect state-action-value pairs $(s,a,v)$ from episodes, where each value 

$$v = \hat{Q}_{em}(s,a) = \sum_{i=0}^T (\gamma)^i \cdot r_{t+i} | s_t =s, a_t = a$$ 

\noindent is the cumulative discounted reward obtained after taking action $a$ in state $s$ (for readability we often omit the dependence of $v$ on $s$ and $a$ when it is clear from the context). For each collected $(s,a,v)$, we then find its closest state neighbour $s^c$ in the current memory. The new state-action-value tuple is directly added into the CEC memory if the state is far away from its closest neighbor. Thereby, if the CEC memory does not contain any information about the area near the incoming state, we directly add it into the memory. 

When its closest neighbor is within a certain distance threshold $d$, and its corresponding value is smaller than the value of the incoming state, we replace the previous state-action-value tuple with the new one. The intuition is that when the closest neighbor is close enough, we assume that the CEC memory has information about the area near the new state. However, if the stored value is worse than the one of the new state-action-value tuple, we do overwrite it since we now have information about a better action in the specific state region. The update rule can be expressed as:
\begin{flalign}
 \text{EM}(s^c, a^c, v^c) \gets
    \begin{cases}
      (s, a, v) & \text{if } f(s,s^c)< d \\ & \text{ and } v>v^c, \\
      (s^c, a^c, v^c)  & \text{otherwise} \\
    \end{cases}
    \label{eq:update}
\end{flalign}
\noindent Here, superscript $c$ indicates the closest neighbour in the buffer, superscript $l$ indicates the least recently updated entry in the buffer, $\text{EM}()$ refers to the slot of the specific entry in the memory buffer, $f()$ specifies a distance measure and $d$ specifies a distance threshold. 
Once the buffer has filled up, when a new coming state is out of the distance threshold of its closest state, the least updated tuple will be thrown away and the new coming tuple will be added, otherwise, it will follow the aforementioned update rule:
\begin{flalign}
 \text{EM}(s^l, a^l, v^l) \gets
    \begin{cases}
      (s, a, v) & \text{if $f(s,s^c) > d$}, \\
      (s^l, a^l, v^l) & \text{otherwise}
    \end{cases} \nonumber \\
    \label{eq:fill_add}
\end{flalign}

\noindent Here, superscript $l$ indicates the least recently updated entry in the buffer.

\paragraph{Action selection} When a state $s$ is encountered, a corresponding action $a$ is selected based on information stored in the CEC memory. We first find the $k$ nearest-neighbors of state $s$, denoted by $S_k$, then filter out neighbors which are too far away. Especially when the memory is sparse, the closest neighbors could still be very far away and we do not consider these faraway neighbors as `similar' states. Next we let values stored for state-action pairs of the filtered neighbors $S_k$ decide which action should be selected. Higher values indicate better actions, but for exploration purposes we still want to give other actions a chance to be chosen. Therefore, we sample an action from the set $S_k$ proportional to their value estimates, by taking a softmax: 

\begin{equation}
    p(a_s|S_k) = \frac{e^{v_s/\tau}}{\sum\limits_{s' \in S_k}e^{v_{s'}/\tau}}, s \in S_k
    \label{eq:p_actions}
\end{equation}

where subscripts $s$ and $s'$ indicate the state the specific action and value belong to in the buffer, and  $\tau$ is a temperature factor to scale exploration. We now have a sample from the discrete set of $a_i$ points in the buffer, but our true action space is of course continuous. We therefore transform our sample to a continuous distribution by adding Gaussian noise to the sampled action with probability $\epsilon$. During the evaluation, all exploration is turned off and actions are selected by greedily taking the action of the closest neighbour. 

\begin{algorithm}[!b]
   \caption{Continuous Episodic Control (CEC)}
   \label{alg:cec}
\begin{algorithmic}
   %\STATE {\bfseries Input:} data $x_i$, size $m$
   \STATE {\bfseries Initialize:} CEC Memory $M$, environment \texttt{Env}, episode memory $M_{eps}$, distance threshold $d$, filter factor $n$, $k$ for k-nearest-neighbors, discount factor $\gamma$, noise standard deviation $\sigma$, exploration factor $\epsilon$.
   %\STATE Define reward function $R$
   \WHILE{training budget left}
   \STATE $s \gets $ \texttt{Env.reset()} \hfill\COMMENT{Reset the environment.}
   \STATE {\it done} = false
   \WHILE{\textbf{not} {\it done}}
   \STATE // SELECT ACTION
   \STATE $S_k =$ \texttt{knn}$(k, s, n, d)$ \hfill\COMMENT{Find $k$ nearest-neighbors of the state $s$ within threshold $n\times d$.}
   \STATE $a \sim p(a_s|S_k)$
   \hfill \COMMENT{Select actions based on Eq.~\ref{eq:p_actions}}.
   \IF{$\Delta\sim U(0,1)<\epsilon$}
   \STATE $\psi \sim \mathcal{N}(0, \sigma^2)$ 
   \STATE $a \gets a+\psi$
   \ENDIF
   \STATE $s', r, done \gets $ \texttt{Env.step($a$)}
   \STATE Append $[s, a, r]$ to $M_{eps}$
   \STATE $s \gets s'$
   \ENDWHILE
   \STATE // UPDATE
   \STATE $v=0$
   \WHILE{$M_{eps}$ \textbf{not} empty}
   \STATE $s,a,r \gets $ $M_{eps}$.\texttt{pop()}
   \STATE $v \gets r+\gamma \cdot v$
   \STATE Update $M$ using $(s,a,v)$. 
   \hfill \COMMENT{Update CEC memory based on Eq.~(\ref{eq:update}~\ref{eq:fill_add})}.
   \ENDWHILE
\ENDWHILE
\end{algorithmic}
\end{algorithm}

\paragraph{Feature embedding} While dealing with a high-dimensional state space, storing the original state and finding its neighbors is expensive. We therefore utilize random projections to project the original state space into a smaller space. We first sample a random matrix $A\in \mathbb{R}^{D_{ori} \times D_{new}}$ for a standard Gaussian distribution, where $D_{ori}$ and $D_{new}$ are the dimensions of the original and projected state space, respectively. We then (matrix) multiply the original state by $A$ to get an embedded state, i.e. $s_{new}=A \cdot s_{ori}$, where $s_{new}$ and $s_{ori}$ are the embedded and original state, respectively. These transformations will still preserve the relative distance in the original state space. We only use the random projection for \textbf{FetchReach} and \textbf{Safexp-CarGoal} environments, while for other environments the original state is directly stored and queried.

The overall algorithm can be found in Alg.~\ref{alg:cec}. \texttt{knn}$(k,s,n,d)$ is a function that first finds the $k$ nearest neighbors of the state $s$, then filters out states whose distances are larger than $n$ times of the distance threshold $d$.

\section{Experiments}

In this section, we will first give some insights of CEC by using simple toy examples. Then we scale up to more complex continuous control tasks and compare with state-of-the-art RL and memory-augmented RL methods. Dimensions of environments we used for this work are shown in Tab.~\ref{tab:d_envs}. Although \textbf{Safexp-PointGoal} and \textbf{Safexp-CarGoal} have the same action space (two actuators, one for turning and one for moving), in \textbf{Safexp-CarGoal} both turning and moving require coordinating both of the actuators, which is more complex. Every experiment is averaged over 5 independent runs and standard errors are plotted as well. 
\begin{table}[!htb]
    \centering
    \caption{Dimensions of state spaces and action spaces of different environments.}
    \begin{tabular}{|c|c|c|}
        \hline
         Env & State Space & Action Space \\
         \hline
         GrowingTree & 1 & 1 \\
         \hline
         MountainCarContinuous & 2 & 1 \\
         \hline
         PointUMaze & 4 & 2 \\
         \hline
         Point4Rooms & 4 & 2 \\
         \hline
         Safexp-PointGoal & 24 & 2 \\
         \hline
         Safexp-CarGoal & 36 & 2 \\
         \hline
         FetchReach & 13 & 4  \\
         \hline
    \end{tabular}
    \label{tab:d_envs}
\end{table}
\subsection{Toy Examples}
\begin{figure}[!htb]
    \centering
    \includegraphics[scale=0.29]{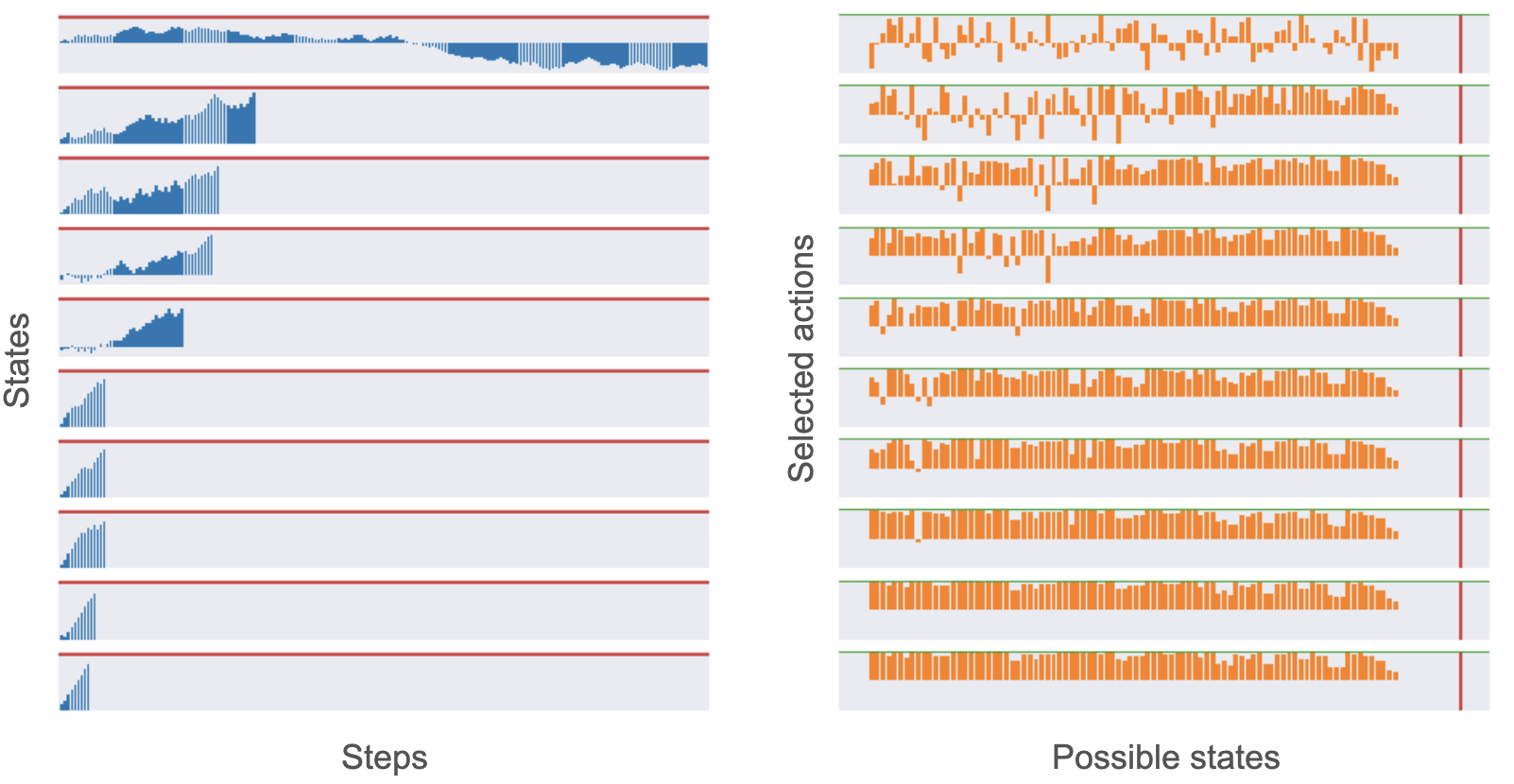}
    \caption{Evaluation on the toy example. We train the CEC agent on the toy example for 100 k steps and evaluate it every 10k steps. Training are from top to bottom. \textbf{Left}: the performance during the evaluation. It gradually takes shorter episode to reach the goal. x-axis is taken steps during the evaluation episode and y-axis is the state (the height of the current `tree'). Red horizontal lines are targets we set. \textbf{Right}: actions selected by the trained CEC agent. x-axis represents different states while y-axis represents actions. Red horizontal lines are final target states and green vertical lines are optimal actions (0.1).}
    \label{fig:toy_results}
\end{figure}
In order to first get an overview and more insights into the method, we create a toy example called GrowingTree. In this environment, the agent needs to `grow' the `tree' to the given target height. More details can be found in Tab.~\ref{tab:details_toy}. The continuous action space ranges from -0.1 to 0.1, which means the agent can chose to `cut' ($<0$) the `tree' or `grow' ($>0$) the `tree'. The observation for the agent is the height of the current `tree' and the height changes every step by simply adding the action to the previous observation. The optimal policy for the agent to get the final reward is to take 0.1 as the action every step whatever the height of the tree is, which causes the `tree' to grow linearly. 
\begin{table}[!tb]
    \centering
    \caption{Details about GrowingTree environment. The agent only gets a final reward when it successfully reaches the given goal (with a distance tolerance).}
    \begin{tabular}{|c|c|}
        \hline
        state space & [-2, 2] \\
        \hline
        action space & [-0.1, 0.1] \\
        \hline
        maximum steps & 200 \\
        \hline
        goal & 1.0 \\
        \hline
        final reward & 1.0 \\
        \hline
        distance tolerance & 0.1 \\
        \hline
        start height & 0 \\
        \hline
    \end{tabular}
    \label{tab:details_toy}
\end{table}

\begin{figure*}[!thb]
    \centering
    \includegraphics[scale=0.5]{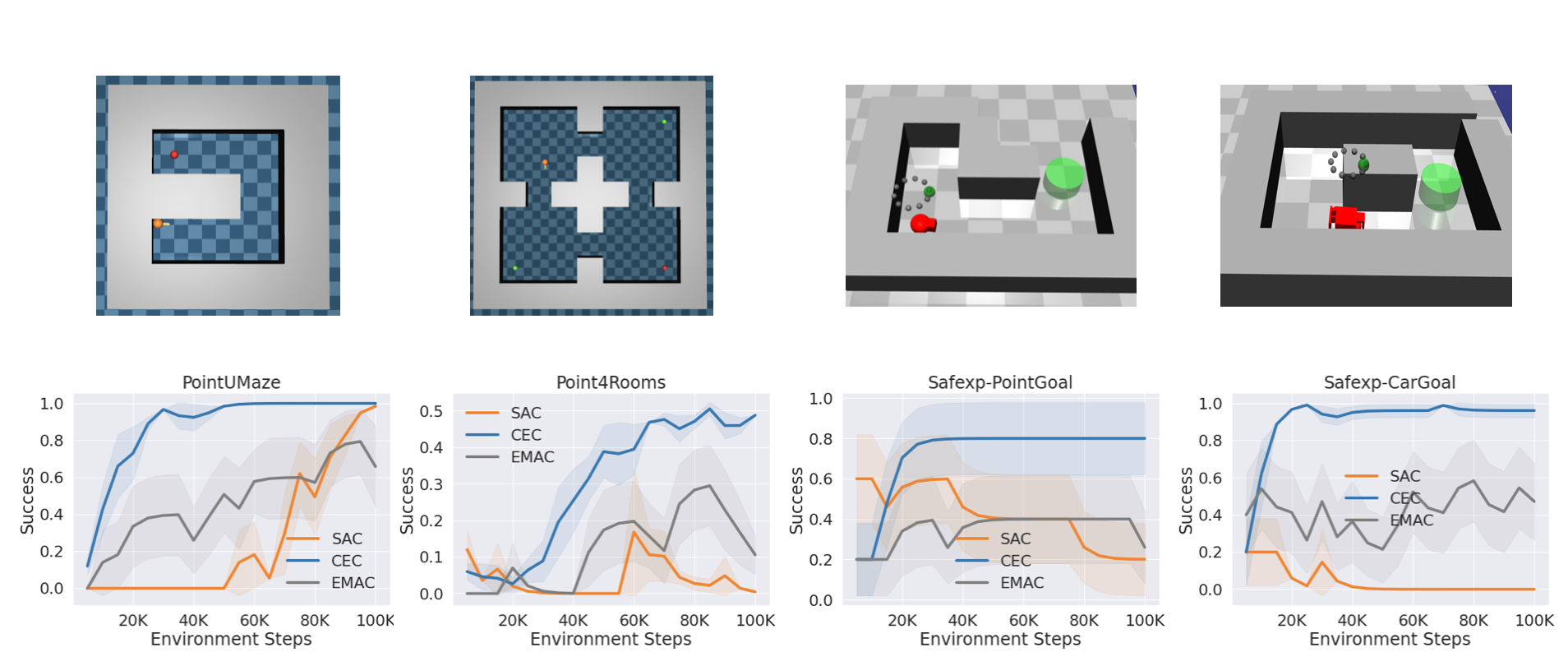}
    \caption{Four continuous navigation environments we used in this work and their corresponding results. From left to right, they are `PointUMaze', `Point4Rooms', `Safexp-PointGoal' and `Safexp-CarGoal', respectively. We see the proposed method CEC (blue lines) outperform SAC (orange lines) and EMAC (grey lines) in all these four tasks. Results are averaged over five independent runs and the shaded area represents standard errors.}
    \label{fig:envs_res}
\end{figure*}

\begin{figure}[!t]
    \centering
    \includegraphics[scale=0.5]{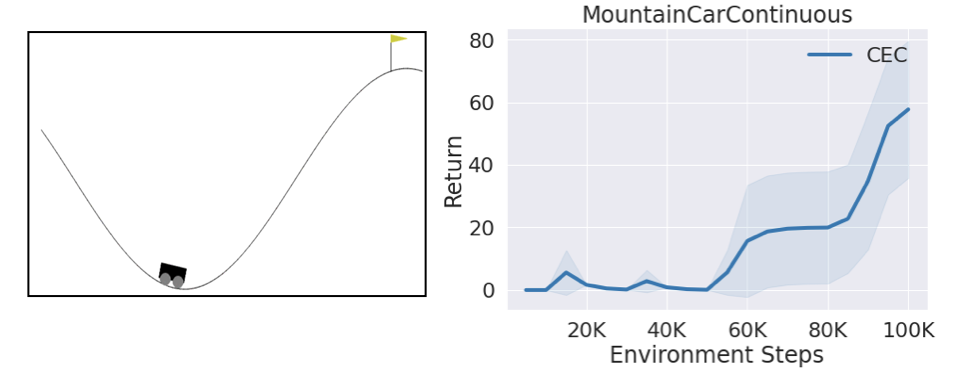}
    \caption{MountainCarContinuous environment and learning curve of the CEC agent on it. The agent gradually learns to solve the problem.}
    \label{fig:mc}
\end{figure}

We train the CEC agent on the toy example, with results shown in Fig.~\ref{fig:toy_results}. Every row of the figure shows an evaluation of the CEC agent after another $10k$ training steps, from top to bottom. The $n^{th}$ row therefore represents the evaluation results after $n \times 10k$ training steps. The left part of the figure shows the heights of the tree during the evaluation episode. If the optimal policy is learned by the agent, the height of the tree will increase quickly with the number of steps (x-axis) increases. The right figure shows the actions the CEC agent takes for every possible state (we only evaluate states that lie within [0,1]). The optimal action in every state is (near) 0.1.

In the top rows, the agent still chooses to `cut' the `tree' sometimes, but the overall trend of the `tree' is still growing. Although the solution is not the optimal, the agent is able to solve the problem successfully. In the middle rows, the agent can solve the task fairly quickly. However, if we look at the middle rows of the right figure, the chosen action for every possible state is still far from the optimal and a few of them are still smaller than 0. Gradually, selected actions of all possible states are getting closer and closer to the optimal values (bottom rows).

Through this simple example, we see that our principle ``similar states should have similar actions" does work and can indeed quickly solve the problem with a non-optimal solution, while it may also discover a near-optimal solution given more time. 

We next test our method in a simple classical RL task, \textbf{MountainCarContinuous}. We use the sparse reward version of the task, where the agent only gets a final reward ($+100$) once it brings the car successfully to the flag on the top of the mountain. Since the initial position of the car is randomly generated near the bottom of the mountain, the agent also needs to deal with proper generalization. The learning curve of CEC on this task is shown in Fig.~\ref{fig:mc}. The CEC agent gradually learns to solve the task, which gives a first indication that our method can indeed be used for basic RL problems that require generalization. 
\begin{figure*}[!hbt]
    \centering
    \includegraphics[scale=0.5]{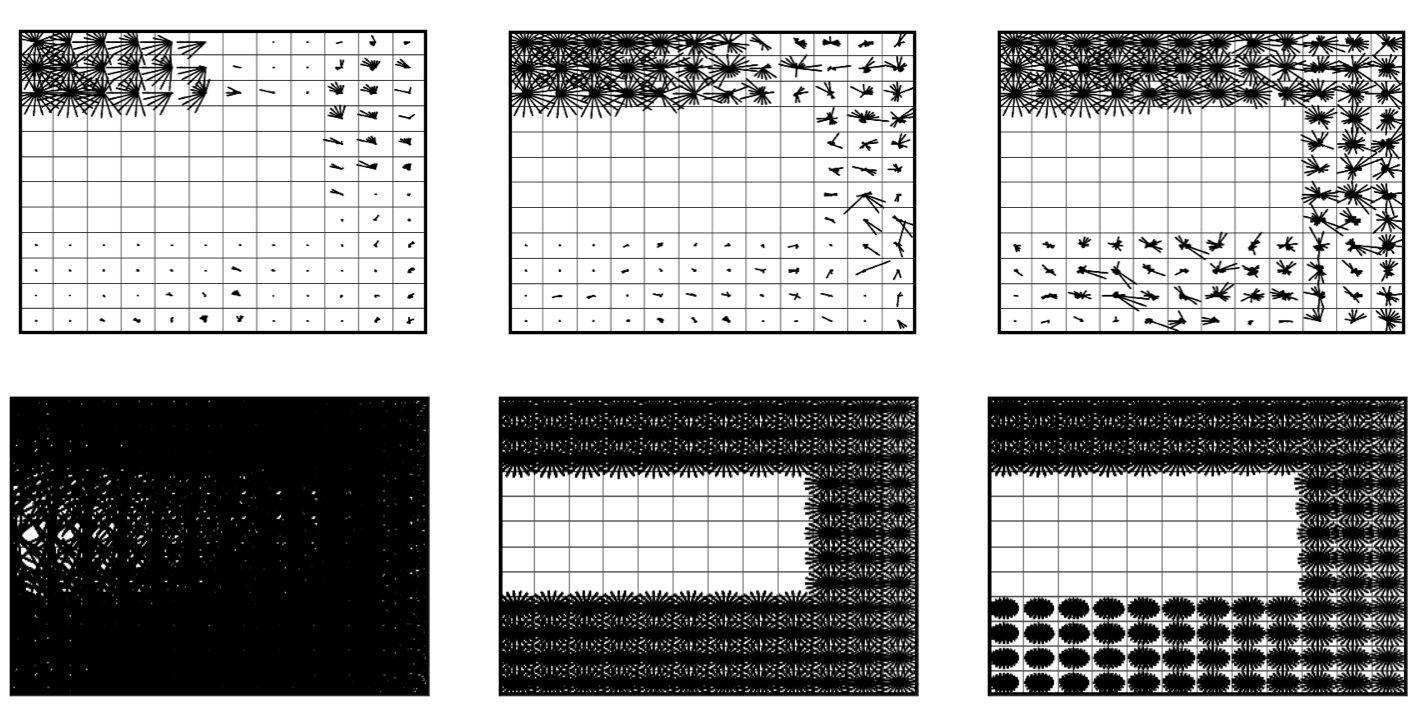}
    \caption{State values produced by CEC (top) and SAC (bottom) during the training on PointUMaze. Figures from left to right are after 5k, 50k, and 100k steps of training, respectively. Since the state space is continuous, we discretize the location information into $12\times 12$ grids and the orientation information into $20$ directions (represent as different directed arrows). Longer arrows indicate larger state-action values, and SAC outputs very large values at the beginning which is not rational while CEC has larger values for the area near the goal.}
    \label{fig:values}
\end{figure*}
\subsection{Continuous Navigation Tasks}
We next move to more complicated test environments, in which we will also compare CEC to two baselines: the model-free deep RL method SAC~\cite{DBLP:conf/icml/HaarnojaZAL18} and a memory-augmented deep RL method EMAC~\cite{ijcai2021-365} which uses EM for learning targets. We test these in four continuous Mujoco navigation tasks, shown in Fig.~\ref{fig:envs_res}. They are PointUMaze, Point4Rooms, Safexp-PointGoal, and Safexp-CarGoal, respectively. These tasks are all sparse-reward problems, in which the agent only gets a final reward when the task is successfully completed. Locations of both goals and agents are initialized with small randomness every episode. Hyper-parameters we are using for CEC on these tasks can be found in Tab~\ref{tab:para}.
\begin{table}[]
    \centering
    \caption{Hyper-parameters of CEC we are using for continuous navigation tasks. Notations can be found in Alg.~\ref{alg:cec}.}
    \begin{tabular}{|c|c|c|c|c|c|}
        \hline
         Env & $k$ & $\tau$ & $\sigma$ & $n$ & $d$ \\
         \hline
         PointUMaze & 5 & 0.1 & 0.3 & 1 & 0.1 \\
         \hline
         Point4Rooms & 5 & 1 & 0.3 & 3 & 0.1 \\
         \hline
         PointGoal & 5 & 0.1 & 0.1 & 1 & 0.1 \\
         \hline
         CarGoal & 1 & 10 & 0.1 & 3 & 0.1 \\
         \hline
    \end{tabular}
    \label{tab:para}
\end{table}

Results are shown in Fig.~\ref{fig:envs_res}. CEC agents can quickly solve the task and outperform SAC and EMAC agents in all environments. In these four navigation tasks, there are explicit bottlenecks and CEC can quickly remember the discovered solution by only discovering it once. In contrast, the deep RL methods need to discover the solution multiple times in order to back-propagate reward signals for learning, and this rediscovery is time-consuming due to the bottlenecks in the tasks.

We also visualize state-action values of CEC and SAC during the training for \textbf{PointUMaze}, which is shown in Fig.~\ref{fig:values}. The agent always starts from the bottom-left corner and the goal is always located at the top-left corner. Both of these locations are initialized with small randomness. Since the state space of the environment is 3 dimensional ($x$ location of the agent, $y$ location of the agent, orientation of the agent) and continuous, we discretize $x$, $y$ locations as a grid world ($12 \times 12$) and orientations as a group of different directed arrows (20 different directions). The length of the arrow represents the state-action value of the state that is jointly indicated by the arrow and the position of the cell and the action for that specific state (only one action is maintained for each state). The longer the arrow is, the larger the state-action value is. From Fig.~\ref{fig:values}, we see that in the beginning the SAC agent strongly overestimates all state-actions values, due to the random initialization(arrows went across the whole state space, meaning all state-action values are extremely large). On the contrary, the CEC agent obtains reasonable estimates more quickly, also identifying larger value estimates near the goal region. While training progresses, both agents gradually improve their estimates to better propagate high-value estimates towards the start region (i.e., identifying a good policy). Both agents eventually solve the task, but CEC was much faster, as can be seen in Fig.~\ref{fig:envs_res} as well. 

\subsection{Robotics Control Task}
\begin{figure}[!tbh]
    \centering
    \includegraphics[scale=0.55]{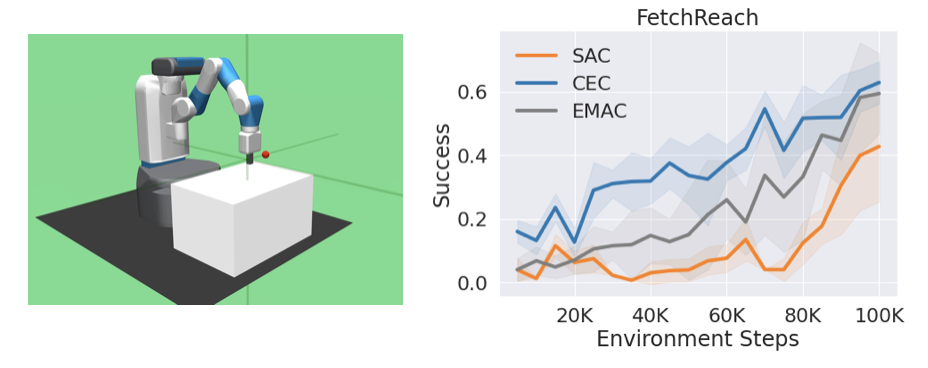}
    \caption{FetchReach environment and performance of different agents. CEC still outperforms other RL agents but the benefits are not as large as the ones in previous navigation tasks.}
    \label{fig:fetchreach}
\end{figure}

We finally investigate our method on a more complicated robotics control task: \textbf{FetchReach}. The goal of the task is to control the robotics arm to reach the red point, which is reset to a randomly sampled position every episode. The environment and its associated learning curve are shown in Fig.~\ref{fig:fetchreach}. We use $k=5, \tau=1, \sigma=0.1, n=3, d=0.5$ for CEC. We see that CEC still outperforms the state-of-the-art baseline methods, although the difference is less pronounced compared to the previous navigation tasks. We attribute this to the lack of bottlenecks in this task. CEC is probably most useful when a task has some narrow passages, which are challenging from an exploration point of view, and in which CEC can quickly latch onto a few successful trials. Nevertheless, even in a robotics task with no bottleneck characteristics, CEC performs on par with current state-of-the-art model-free baselines. 

\section{Conclusion and Future Work}
In this work, we propose a new episodic control method called continuous episodic control (CEC) which can handle tasks with a continuous action space. It is the first work that uses episodic memory for continuous action selection. By following the principle that ``similar states should have similar actions", CEC agents outperform the state-of-the-art RL method SAC and memory-augmented RL method EMAC in various sparse-reward continuous control environments. Although RL methods might outperform episodic control in the long run, sometimes we do prefer a quick and cheap solution to the problem. Several experiments show CEC has strong performance on continuous navigation tasks, especially when there are bottlenecks in the environment.

Although non-parametric methods (like episodic control) require increasing memory with new data, this can be eliminated by employing a throw-away mechanism that only stores important data. In fact, experimental results show that episodic control is able to scale up to Atari games~\cite{DBLP:journals/nature/SilverHMGSDSAPL16}, such as Ms.Pac-Man, Space Invaders, and Frostbite, where pixel images are used as observations. However, episodic control also has its limitations. For example, using Monte-Carlo returns as its state-action values has high variance, and also it will not be able to learn the optimal solution in stochastic environments by definition. But we believe sometimes we do prefer a quick (might be sub-optimal) solution.
%Since episodic control approaches (including CEC) use Monte-Carlo returns as their state-action values, which have very high variance, it is not well-suited for non-episodic tasks such as Mujoco Swimmer or HalfCheetah, etc. Meanwhile, if the environment is stochastic, episodic control approaches will be problematic by definition. Another limitation is that episodic control approaches usually enjoy generalization from KNN over unlearnable feature space, although this problem can be eliminated by using pre-trained features (trained using VAE~\cite{kingma2013auto} to predict next states given current states and actions), it does not make use of the association of reward structure and feature space~\cite{liureturn} compare to training features end-to-end using reward signals in RL. In light of the aforementioned limitations of episodic control approaches, we suggest researchers who are interested in this topic carefully consider these drawbacks before diving into them. 

In the future, it would be very interesting to investigate better feature embedding methods (such as using latent features of a pre-trained VAE~\cite{kingma2013auto}) and exploration strategies that work well with CEC. In addition, we could also look at methods to improve the nearest neighbour search, for example, based on clustering. Finally, we believe there is potential to combine episodic control and parametric reinforcement learning methods to distill strengths from both sides and then end up with a more powerful agent.

\bibliographystyle{IEEEtran}
\bibliography{IEEEabrv,ref}

\end{document}